\documentclass{article}

\usepackage[preprint]{neurips_2021}

\usepackage[utf8]{inputenc} 
\usepackage[T1]{fontenc}    
\usepackage{hyperref}       
\usepackage{url}            
\usepackage{booktabs}       
\usepackage{amsfonts}       
\usepackage{nicefrac}       
\usepackage{microtype}      
\usepackage{xcolor}         

\usepackage{amsmath,amsfonts,bm}

\def\thetagenie{\hat{\theta}(\mathcal{D}_N;x,y)}
\def\thetagenietag{\hat{\theta}(\mathcal{D}_N;x,y')}
\def\probthetagenie{p_{\thetagenie}(y|x)}
\def\probthetagenietag{p_{\thetagenietag}(y'|x)}

\newcommand{\norm}[1]{\left\lVert#1\right\rVert}
\def\Theoref#1{Theorem~\ref{#1}}
\def\appref#1{appendix~\ref{#1}}

\def\lemmaref#1{lemma~\ref{#1}}
\def\Lemmaref#1{Lemma~\ref{#1}}
\def\tableref#1{table~\ref{#1}}

\def\figref#1{figure~\ref{#1}}
\def\Figref#1{Figure~\ref{#1}}

\def\secref#1{section~\ref{#1}}

\def\eqref#1{(\ref{#1})}

\def\1{\bm{1}}

\DeclareMathAlphabet{\mathsfit}{\encodingdefault}{\sfdefault}{m}{sl}
\SetMathAlphabet{\mathsfit}{bold}{\encodingdefault}{\sfdefault}{bx}{n}

\newcommand{\normltwo}{L^2}

\DeclareMathOperator*{\argmin}{arg\,min}

\usepackage{url}
\usepackage{amsthm}
\usepackage{amsmath,amssymb}
\usepackage{multirow}
\usepackage{wrapfig}
\usepackage{graphicx}
\usepackage[toc,page]{appendix}
\usepackage{placeins}
\usepackage{array}
\usepackage{lipsum} 
\usepackage{enumitem}

\usepackage{caption}        
\usepackage{subcaption} 

\newtheorem{theorem}{Theorem}
\newtheorem{lemma}{Lemma}

\newcommand{\ignore}[1]{}

\usepackage{soul}
\newcommand{\minisection}[1]{\vspace{2mm}\noindent{\textbf{#1.}}}

\title{Distribution Free Uncertainty for the Minimum Norm Solution of Over-parameterized Linear Regression}

\author{%
    Koby Bibas \\
    School of Electrical Engineering \\
    Tel Aviv University\\
    \texttt{kobybibas@gmail.com} \\
    \And
    Meir Feder\\
    School of Electrical Engineering \\
    Tel Aviv University\\
    \texttt{ meir@eng.tau.ac.il} \\
}

\begin{document}

\maketitle

\begin{abstract}
A fundamental principle of learning theory is that there is a trade-off between the complexity of a prediction rule and its ability to generalize. Modern machine learning models do not obey this paradigm: They produce an accurate prediction even with a perfect fit to the training set. We investigate over-parameterized linear regression models focusing on the minimum norm solution: This is the solution with the minimal norm that attains a perfect fit to the training set. We utilize the recently proposed predictive normalized maximum likelihood (pNML) learner which is the min-max regret solution for the distribution-free setting. We derive an upper bound of this min-max regret which is associated with the prediction uncertainty. We show that if the test sample lies mostly in a subspace spanned by the eigenvectors associated with the large eigenvalues of the empirical correlation matrix of the training data, the model generalizes despite its over-parameterized nature. We demonstrate the use of the pNML regret as a point-wise learnability measure on synthetic data and successfully observe the double-decent phenomenon of the over-parameterized models on UCI datasets.
\end{abstract}

\section{Introduction}
\label{sec:introduction}

Classic learning theory argues that complex models tend to overfit their training set, thus generalizing poorly to unseen ones~\citep{bartlett2020benign,hastie2001friedman,kaufman2019balancing}.
This assumption is challenged by modern learning models such as deep neural networks (DNN) which operate well even with a perfect fit to the training set~\citep{DBLP:conf/iclr/ZhangBHRV17}.
Motivated by this phenomenon, we consider when a perfect fit to the training set is compatible with an accurate prediction, i.e., when a small \textit{generalization error} is achieved.

We examine \textit{over-parameterized} linear regression, where the number of the learnable parameters is larger than the training set size.
We focus on the \textit{minimum norm} (MN) solution. This solution has the following unique property: It is the solution with the minimal norm that attains a perfect fit on the training set.
Recent work show that the MN solution generalizes well in the over-parameterized regime and approximated its generalization error~\citep{liang2020just,ma2019generalization,hastie2019surprises,shah2018minimum,belkin2018overfitting}. However, they assume some probabilistic connection between the training and test which may not be valid in a real-life scenario.

We exploit the \textit{individual setting} framework~\citep{merhav1998universal}. 
In this framework there is no assumption on how the training set and the tested sample are generated, nor about their probabilistic relationship. Both are specific individual values without any distribution assumption.
The absence of assumption implies that the individual setting is the most general framework and so the result holds for a wide range of scenarios.
The common approach in the individual setting is to select a learner that can compete with a reference learner, a \textit{genie}. This genie knows the test label value, yet is constrained to use an explanation from a class of possible models or \textit{hypothesis set}. 
The \textit{regret} is defined as the logloss difference between a learner and this genie and is associated with the generalization error: 
When the regret is small the prediction is similar to the genie and thus it can be trusted.

The pNML learner was recently proposed as the min-max regret solution in the individual setting~\citep{fogel2018universal}, where the minimum is over the learner choice and the maximum is for the test label value.
Previous work stated the pNML learner for \textit{under-parameterized} hypothesis sets, where the number of the learnable parameters is smaller than the training set size.
These work dealt with 1D barrier~\citep{fogel2018universal}, linear regression~\citep{bibas2019new}, and the last layer of DNN~\citep{bibas2019deep}.
Using a large hypothesis set in the pNML procedure produces the maximal regret for every test sample therefore is not informative in providing generalization error.

We derive an upper bound of the pNML regret for over-parameterized linear regression.
We design the hypothesis set to contain hypotheses that have a norm that is not larger than the MN norm.
By utilizing this hypothesis set, the pNML prediction equals the MN solution. Thus the derived upper bound of the regret can be used as the generalization error of the MN solution.
We show that if the test vector resides in a subspace spanned by the eigenvectors associated with the large eigenvalues of the empirical correlation matrix of the training data, linear regression can generalize despite its over-parameterized nature.
In addition, we present a recursive formulation of the norm of the MN solution. 
We demonstrate the case where a small deviation from the MN solution prediction increases the model norm significantly, which implies high confidence in the MN prediction.

To summarize, we make the following contributions.
\begin{itemize}
    \item \textbf{Designing the norm constrained hypothesis set.} Introducing the norm constrained hypothesis set for over-parameterized linear regression. Utilizing this set, we create a pNML learner that has a meaningful regret and a prediction that equals the MN solution.
    \item \textbf{Upper bounding the pNML regret.}  Deriving an analytical upper bound of the pNML regret, which is associated with the generalization error. We demonstrate what are the characteristics of the test data for which the regret is small.
    \item \textbf{Deriving a recursive formulation for the norm of the MN solution.} We present a recursive formula for the norm of the MN solution. We show what are the properties of the test data for which a small deviation from the MN prediction increases significantly the norm of the MN solution. In this situation, the MN prediction is considered reliable, as it has a significantly smaller norm than the other predictors that fit the training data.
\end{itemize}

The presented results hold for nearly all settings since the pNML is the min-max solution of the individual setting in which there is no assumption on a probabilistic connection between the training set and the test sample (distribution-free).
In addition, we demonstrate the use of the pNML regret as a confidence measure in a simulation of fitting trigonometric polynomials to synthetic data. Also, we show that the empirically calculated regret and its upper bound are correlated with the test error double-decent phenomenon on sets from the UCI repository~\citep{Dua:2019}.

\section{Related work} 
\label{sec:related_work}



\minisection{Over-parameterized linear regression}
A popular approach to deal with over-parameterized models is to find the optimal regularization term for the ridge regression model class.
\citet{nakkiran2020optimal} showed that models with optimally-tuned regularization achieve monotonic test performance as growing the model size.
\citet{dwivedi2020revisiting} used the minimum description length principle to quantify the linear model complexity and to find the optimal regularization term.
\citet{kobak2020optimal} claimed that for over-parameterized linear regression the optimal ridge penalty can be negative.

Several studies suggested that for linear regression the generalization is proportional to the model norm~\citep{kakade2009complexity,shamir2015sample,ma2019generalization}.
\citet{muthukumar2020harmless} showed that the generalization error decays to zero with the number of features.
\citet{tsigler2020benign} provided non-asymptotic generalization bounds for over-parameterized ridge regression.
\citet{nichani2020deeper} analyzed the effect of increasing the depth of linear networks on the test error using the MN solution.
\citet{hastie2019surprises} provided a non-asymptotic approximation of the generalization error for the over-parameterized region.
In addition, several authors argued that the MN solution captures the basic behavior of DNN~\citep{allen2019convergence,gunasekar2018implicit}.

However, all mentioned work assume some probability distribution on the training and testing sets or the learnable parameters. This assumption may not apply in a real-life scenario. Moreover, they do not consider the specific test input thus do not provide a point-wise generalization error.

\minisection{The pNML learner}
The pNML learner is the min-max regret solution of the supervised batch learning in the individual setting~\citep{fogel2018universal}. For sequential prediction this learner was suggested by \citet{roos2008sequentially} and was termed the conditional normalized maximum likelihood (CNML).
It follows the normalized maximum likelihood method~\citep{shtar1987universal}.

Several work deal with obtaining the pNML learner for different hypothesis sets:
\citet{rosas2020learning} proposed an NML based decision strategy for supervised classification problems and showed that it attains heuristic PAC learning.
\citet{bibas2019new} showed the pNML solution for the under-parameterized linear regression case. However, as similar generalization measures, when using an over-parameterized hypothesis set, their derived regret becomes infinite and therefore cannot be used.

For the DNN hypothesis set,
\citet{bibas2019deep} estimated the pNML distribution with DNN by fine-tuning the last layers of the network on every test input and label combination.
\citet{zhou2020amortized} suggested a way to accelerate the pNML computation in DNN by using approximate Bayesian inference techniques to produce a tractable approximation of the pNML distribution.

\section{Notation and preliminaries}
\label{sec:preliminaries}

In the supervised machine learning scenario, a training set consisting of $N$ pairs of examples is given
\begin{equation}
\mathcal{D}_N = \{(x_n,y_n)\}_{n=1}^{N}, \quad x_n \in R^{M \times 1}, \quad y_n \in R,
\end{equation}
where $x_n$ is the $n$-th data instance and $y_n$ is its corresponding label.
The goal of a learner is to predict the unknown label $y$ given a new test data $x$ by assigning a probability distribution $q(\cdot|x)$ to the unknown label. 
The performance is evaluated using the logloss function
\begin{equation}
\ell(q;x,y) = -\log {q(y|x)}.
\end{equation}
For the problem to be well-posed, we must make further assumptions on the class of possible models or the hypothesis set that is used in order to find the relation between $x$ and $y$.
Denote $\Theta$ as a general index set, this class is a set of conditional probability distributions 
\begin{equation} \label{eq:hypotesis_set}
P_\Theta = \{p_\theta(y|x)\ | \;\;\theta\in\Theta\}.
\end{equation}
The \textit{empirical risk minimizer} (ERM) is the learner from this hypothesis set that attains the minimal log-loss on the training set.

\minisection{The individual setting}
An additional required assumption is on the generation of the data and the labels. 
We consider the individual setting~\citep{fogel2018universal,merhav1998universal},
where the data and labels, both in the training and test, are specific individual quantities: We do not assume a probabilistic relationship between them. The labels may be assigned in an adversarial manner.

\minisection{The genie} In the individual setting, the goal is to compete with a reference learner, a genie. This genie has the following properties: (i) knows the desired test label value, (ii) is restricted to use a model from the given hypothesis set $P_\Theta$, and (iii) does not know which of the samples is the test. 
This reference learner then chooses a model that attains the minimum loss over the training set and the test sample
\begin{equation} \label{eq:genie_pnml} 
\hat{\theta}(\mathcal{D}_N;x,y)  = \arg\min_{\theta \in \Theta} \left[
\ell(p_\theta;x,y) + \sum_{n=1}^N \ell(p_\theta;x_n,y_n)
\right].
\end{equation}
The regret is the logloss difference between a learner $q$ and this genie:
\begin{equation} \label{eq:regret}
R(q;\mathcal{D}_N;x,y) = - \log q(y|x) - \left[- \log p_{\thetagenie}(y|x)\right].
\end{equation}

\begin{theorem}[\citet{fogel2018universal}]
\label{theroem:pnml}
The universal learner, denoted as the pNML, minimizes the worst case test label objective
\begin{equation} \label{eq:minmax_prob}
\Gamma = \min_q \max_{y \in \mathcal{Y}} R(q;\mathcal{D}_N;x,y).
\end{equation}
The pNML probability assignment and regret are
\begin{equation} \label{eq:pNML}
q_{\mbox{\tiny{pNML}}}(y|x)= \frac{\probthetagenie}{\sum_{y' \in \mathcal{Y}} \probthetagenietag} ,
\quad \Gamma = \log \sum_{y' \in \mathcal{Y}} \probthetagenietag.
\end{equation}
\end{theorem}
\begin{proof}
The regret is equal for all choices of $y$. If we consider a different probability assignment, it should assign a smaller probability for at least one of the outcomes. If the true label is one of those outcomes it will lead to a higher regret. For more information see \citet{fogel2018universal}.
\end{proof}
The pNML regret is associated with the model complexity~\citep{zhang2012model}. This complexity measure formalizes the intuition that a model that fits almost every data pattern very well would be much more complex than a model that provides a relatively good fit to a small set of data.
Thus, the pNML incorporates a trade-off between goodness of fit and model complexity as measured by the regret.

\citet{bibas2019new} derived the pNML regret for under-parameterized linear regression.
\begin{theorem}[\citet{bibas2019new}] \label{theorem:under_parameterized_pnml}
Denote the data matrix and label vector as
\begin{equation} \label{eq:trainset_matrix}
X_N = 
\begin{bmatrix}
x_1 & x_2 & \dots & x_N
\end{bmatrix}^\top \in \mathcal{R}^{N \times M},
\quad
Y_N = \begin{bmatrix} y_1 & y_2 & \dots & y_N \end{bmatrix}^\top \in \mathcal{R}^{N \times 1},
\end{equation}
and let $u_m$ and $h_m$ be the $m$-th eigenvector and eigenvalues of the training set data matrix.
Assuming $X_N^\top X_N$ is invariable ($M \leq N$), the pNML regret and normalization factor are
\begin{equation}
\Gamma = \log K_0, \qquad
K_0 = 1 + \frac{1}{N} \sum_{m=1}^{M} \frac{\left(x^\top u_m\right)^2 }{h_m^2}.
\end{equation}
\end{theorem}
This result deals with under-parameterized linear regression models. It shows that if the test sample $x$ lies in the subspace spanned by the eigenvectors with large eigenvalues, the corresponding regret is low. In this situation the model prediction is similar to the genie's and can be trusted.

\minisection{The minimum norm solution}
The Moore-Penrose inverse of $X_N$ is~\citep{ben2003generalized}
\begin{equation} \label{eq:pseudo-inverse}
X_N^+ = 
\begin{cases} 
    (X_N^\top X_N)^{-1} X_N^\top & \textit{Rank}(X_N^\top X_N) = M, \\
    X_N^\top (X_N X_N^\top)^{-1} & \textit{otherwise}.   
\end{cases}
\end{equation}
In over-parameterization, the MN solution is the solution that attains a perfect fit to the training set and has the lowest norm
\begin{equation} \label{eq:mn_solution}
\theta_N^* = X_N^+ Y_N.
\end{equation}
Denote the ridge regression solution as 
\begin{equation} \label{eq:ridge}
\theta_N^\lambda \triangleq \left(X_N^\top X_N  + \lambda I \right)^{-1} X_N^\top Y_N 
\end{equation} 
where $\lambda$ is the regularization term, it can be shown that $\lim_{\lambda \xrightarrow{} 0} \theta_N^\lambda = \theta_N^*$~\citep{zhou2002variants}.

\section{The norm of the minimum norm solution} 
\label{sec:mn_norm}

We present the behavior of the norm of the MN solution for an over-parameterized linear regression model. 
We show the properties of the test sample for which the MN solution prediction can be trusted.
In~\secref{sec:regret_upper_bound}, we use this result to upper bound the regret.

\begin{theorem}  \label{theorem:mn_norm}
Denote the projection of the test sample onto the orthogonal subspace of the training data empirical correlation matrix as 
\begin{equation}
x_\bot \triangleq \left[I - X_N^+ X_N \right] x,
\end{equation}
the norm of the MN solution based on the training set $\mathcal{D}_N$ and the test sample $(x,y)$ is given by 
\begin{equation}  \label{eq:mn_solution_norm}
\norm{\theta_{N+1}^*}^2 = \norm{\theta_N^*}^2 + \frac{1}{\norm{x_\bot}^2}(y-x^\top \theta_N^*)^2.
\end{equation}
\end{theorem}
\begin{proof}
The recursive form of the MN solution based on $N+1$ samples is~\citep{zhou2002variants}
\begin{equation}
    \theta_{N+1}^* = \theta_N^* +  x_\bot^{+ \top} (y - x^\top \theta_N^*).
\end{equation}
Computing its norm gives the desired results. 
The full derivation is in \appref{appendix:mn_norm_behaviour}.
\end{proof}

If the test sample $x$ lies mostly in the subspace that is spanned by the eigenvectors of the empirical correlation matrix of the training data then $\norm{x_\bot}$ is small: A slight deviation from the MN solution prediction increases significantly the norm of $\theta_{N+1}^*$ (the MN solution that includes the test sample).
On the other hand, if the test sample lies in the orthogonal subspace, $\norm{x_\bot}$ is relatively large and a deviation from the MN solution prediction does not change the norm of $\theta_{N+1}^*$.

If many values of the test label produce MN solution with a low norm, they are all reasonable and therefore none of them can be trusted. On the contrary, if there is just one value of the test label that is associated with MN solution with a low norm, we are confident that it is the right one. 
For confident prediction, we would like that any other prediction will cause a model with high complexity, this is a situation where $\norm{x_\bot}$ is small.
We use this result to upper bound the pNML regret in~\secref{sec:regret_upper_bound}.

\section{The pNML learner for over-parameterized linear regression}
\label{sec:pnml_learner}


\subsection{Formal problem definition} 
\label{sec:problem_def}

For linear regression, we assume a linear relationship between the data and labels
\begin{equation}
y_i = x_i^\top \theta + e_i,
\qquad 1 \leq i \leq N, 
\quad x_n \in R^{M \times 1}, 
\quad y_n \in R.
\end{equation}
$e_i$ is a white noise random variable with a variance of $\sigma^2$.
The test label $y$, conditioned on the test data $x$, has a normal distribution that depends on the learnable parameters
\begin{equation} \label{eq:p_theta}
p_{\theta}(y|x) 
=\frac{1}{\sqrt[]{2\pi\sigma^2}}\exp\left\{-\frac{1}{2\sigma^2}\big(y- x^\top \theta \big)^2\right\}.
\end{equation}
The unknown vector $\theta$ belongs to a set $\Theta$.
A different perspective that corresponds to the individual setting is to assume that $x$ and $y$ are deterministic individual values and the given hypothesis set that the genie can choose from is composed of learners that are defined by \eqref{eq:p_theta}.

Executing the pNML procedure using an over-parameterized hypothesis set would lead to noninformative regret:
Having a large hypothesis set may produce a perfect fit to every test label and therefore the maximal regret.
To reduce the hypothesis set size, we include learners whose $\normltwo$ norm is not larger than the norm of the MN solution
\begin{equation} \label{eq:mn_hypotesis_class}
P_\Theta = \left\{p_\theta(y|x) \ | \ \norm{\theta} \leq \norm{\theta_N^*} \ , \ \theta \in R^{M \times 1} \ \right\}.
\end{equation}
Our goal is to find the pNML regret of \eqref{eq:pNML}, using this hypothesis set as defined in \eqref{eq:mn_hypotesis_class}.

\subsection{The pNML regret upper bound} \label{sec:regret_upper_bound}

We now show an upper bound of the pNML regret using the hypothesis set that contains only learners that have a norm that is not larger than the MN norm.

The genie~\eqref{eq:genie_pnml} that knows the test label value is the solution of the following minimization objective
\begin{equation}
\thetagenie = \argmin_{\theta \in R^{M \times 1}} \left[ \left(y - x^\top \theta \right)^2 + \sum_{n=1}^N \left(y_n - x_n^\top \theta \right)^2\right] \text{ s.t.} \norm{\theta} = \norm{\theta_N^*}.
\end{equation}
With \eqref{eq:ridge}, we write the genie using the recursive least squares formulation~\citep{hayes19969}
\begin{equation} \label{eq:theta_genie_rls}
\thetagenie = \theta_N^\lambda + \frac{\left(X_N^\top X_N + \lambda I\right)^{-1} x}{1 + x^\top \left(X_N^\top X_N + \lambda I \right)^{-1} x} \left(y - x^\top \theta_N^\lambda\right).
\end{equation}
Notice that $\lambda$ depends on the test label $y$ such that the norm constraint is fulfilled.

\begin{lemma} \label{lemma:genie_upper_bound}
The upper bound of the genie probability assignment is
\begin{equation} \label{eq:genie:upper_bound}
\probthetagenie
\leq 
\frac{1}{\sqrt{2\pi\sigma^2}}
\exp\left\{
- \frac{\left(y-x^\top \theta_N^\lambda \right)^2}{2\sigma^2 K_0^2 \left(1 + \frac{\norm{x_\bot}^2}{K_0 \lambda} \right)^2} 
\right\},
\end{equation}
\end{lemma}
where $K_0$ is defined in \Theoref{theorem:under_parameterized_pnml}
and $\lambda$ satisfies the norm constraint
$||\thetagenie|| = \norm{\theta_N^*}$.
\begin{proof}
The genie probability assignment using the recursive formulation~\eqref{eq:theta_genie_rls} is
\begin{equation} \label{eq:prob_genie_rls}
\begin{split}
\probthetagenie &=
\frac{1}{\sqrt{2\pi\sigma^2}}
\exp\left\{
-\frac{1}{2\sigma^2} \left(y - x^\top \thetagenie \right)^2
\right\}
\\ &=
\frac{1}{\sqrt{2\pi\sigma^2}}
\exp\left\{
-\frac{\left(y - x^\top \theta_N^\lambda \right)^2}{2\sigma^2\left[1 + x^\top \left(X_N^\top X_N + \lambda I \right)^{-1} x\right]^2} 
\right\}.
\end{split}
\end{equation}
Let $u_m$ and $h_m$ be the $m$-th eigenvector and eigenvalues of the training set data matrix (using SVD decomposition). 
Assuming over-parameterization $N < M$,
\begin{equation} \label{eq:1_plus_x_P_x_upper_bound}
1 + x^\top \left(X_N^\top X_N + \lambda I \right)^{-1} x = 
1 +
\sum_{m=1}^N \frac{\left( u_m^\top x \right)^2}{h_m^2 + \lambda} 
+
\sum_{m=N+1}^M \frac{\left( u_m^\top x \right)^2}{\lambda}
\leq
K_0 + \frac{1}{\lambda} \norm{x_\bot}^2.
\end{equation}
where we set $\lambda=0$ for $m\leq N$.
Substitute \eqref{eq:1_plus_x_P_x_upper_bound} to \eqref{eq:prob_genie_rls} proves the lemma.
\end{proof}
The genie probability distribution is monotonic decreasing with respect to $\lambda$.

\begin{lemma} \label{lemma:lambda_lower_bound}
The lower bound of the regularization term $\lambda$ that satisfies $||\thetagenie||=\norm{\theta_N^*}$ is
\begin{equation}
\lambda 
\geq
\frac{1}{2}
\frac{\frac{1}{\norm{x_\bot}^2}\left(y - x^\top \theta_N^* \right)^2 }{\theta_N^{*\top} X_N^+ X_N^{+ \top} \theta_N^*
+
\frac{\left(y - x^\top \theta_N^* \right)^2}{||x_{\bot}||^2} 
x^\top X_N^+ X_N^{+ \top} x}.
\end{equation}
\end{lemma}
\begin{proof}
The proof is given in \appref{appendix:lambda_lower_bound}.
\end{proof}
When $y$ equals the MN solution prediction $x^\top \theta_N^*$, the regularization term is zero. As $y$ deviates from the MN solution prediction, the $\lambda$ that is needed to fulfill the norm constraint increases.
This implies that the pNML probability assignment is maximal for the MN prediction and is symmetric around it.

For each possible value of the test label $y'$, we wish to find the learner that satisfies the norm constraint $||\thetagenietag|| = ||\theta_N^*||$
and use in the pNML regret calculation
\begin{equation} \label{eq:overparam_norm_factor}
\Gamma = \log 
\int_{-\infty}^\infty
\frac{1}{\sqrt{2 \pi \sigma^2}} \exp \left\{-
\frac{1}{2\sigma^2} \left(y- x^\top \thetagenietag \right)^2 \right\} dy'.
\end{equation}
\begin{theorem} \label{theorem:regret_upper_bound}
The norm constrained pNML regret upper bound is
\begin{equation} \label{eq:regret_upper_bound}
\Gamma 
\leq
\log \left[
\left(1 + x^\top X_N^+ X_N^{+ ^\top} x \right)
\left(1 + 2\norm{x_\bot}^2 \right)
+ 
3\sqrt[3]{
\frac{1}{\pi \sigma^2}
\norm{x_\bot}^2
\theta_N^{*\top} X_N^+ X_N^{+ \top} \theta_N^*
}
\right]
\end{equation}
\end{theorem}
\begin{proof}
Denote $\delta \geq 0$, we relax the constraint
\begin{equation}
\norm{\theta_N^*}^2 = ||\thetagenietag||^2 \leq (1+\delta) \norm{\theta_N^*}^2
\end{equation}
Using \Theoref{theorem:mn_norm}, a perfect fit is attained when the following constraint is satisfied 
\begin{equation}
|y' - x^\top \theta_N^*| \leq |\Tilde{y}|, \qquad 
\Tilde{y} \triangleq  x^\top \theta_N^* + \sqrt{\delta \norm{x_\bot}^2 \norm{\theta_N^*}^2}.
\end{equation}
We upper bound the regret with the relaxed constraint: For all $y'$s up to $\Tilde{y}$ we get a perfect fit, and for $y'$ from $\Tilde{y}$ to infinity we use the upper bound from \lemmaref{lemma:genie_upper_bound}
\begin{equation}
\Gamma \leq \log \left[  
2 \int_{x^\top \theta_N^*}^{\Tilde{y}} \frac{1}{\sqrt{2 \pi \sigma^2}} dy' \ +
2 \int_{\Tilde{y}}^{\infty} 
\frac{1}{\sqrt{2\pi\sigma^2}}
\exp\left\{
- \frac{\left(y'-x^\top \theta_N^\lambda \right)^2}{2\sigma^2 K_0^2 \left(1 + \frac{\norm{x_\bot}^2}{K_0 \lambda} \right)^2}
\right\}
dy' \right].
\end{equation}
Next, we fix $\lambda$ at the point $\Tilde{y}$. We use the lower bound from \lemmaref{lemma:lambda_lower_bound} to further upper bound the expression.
After integrating, we get the regret that depends on $\delta$.
To get a tight bound, we find $\delta$ that minimizes the regret and that proves the theorem.
The complete derivation is given in \appref{appendix:pnml_regret_upper_bound}.
\end{proof}

In supervised machine learning, the training set is given. We are interested in what conditions the test sample is associated with a low generalization error.
We make the following remarks.
\begin{enumerate}
\item 
The regret is proportional to $x^\top X_N^+ X_N^{+ \top} x $ term.
Denote $u_m$ and $h_m$ as the $m$-th eigenvector and eigenvalue of the training data matrix $X_N$
\begin{equation}
K_0 =  1 + x^\top X_N^+ X_N^{+ ^\top} x = 1 + \frac{1}{N} \sum_{m=1}^{\min \left(M,N\right)} \frac{\left(x^\top u_m\right)^2 }{h_m^2}.
\end{equation}
This term is small when the test sample lies within the subspace spanned by the eigenvectors of the training set empirical correlation matrix that is associated with the large eigenvalues.
Also, $K_0$ decreases when increasing the training set size.
\item 
According to~\eqref{eq:regret_upper_bound}, when $\norm{x_\bot}=0$, the regret upper bound equals the under-parameterized pNML regret $\log K_0$ as in \Theoref{theorem:under_parameterized_pnml}.
\item 
As more energy of the test sample is found in the orthogonal subspace of the training data correlation matrix, $\norm{x_\bot}$ increases and so does the regret.
\item
The pNML regret is proportional to the term 
\begin{equation}
\theta_N^{*\top} X_N^+ X_N^{+ \top} \theta_N^* = \norm{X_N^{+ \top} X_N^{+} Y_N}^2.  
\end{equation}
This term represents the norm of the MN solution. 
This is similar to the works described in \secref{sec:related_work} that show that linear regression generalization is proportional to the model norm.
\item
Increasing $\sigma^2$ reduces the pNML regret. This may relate to the genie: Increasing the noise reduces the genie's performance, which makes the pNML logloss closer to the genie's.
\end{enumerate}

\section{Experiments} 
\label{sec:experimnets}

In this section, we detail the experiments that show the pNML behavior on synthetic and real datasets.
Also, we show that the pNML regret is a good indicator for the double-decent phenomenon.

\begin{figure}[h]
\centering
\begin{subfigure}[t]{0.45\linewidth}
    \includegraphics[width=\columnwidth]{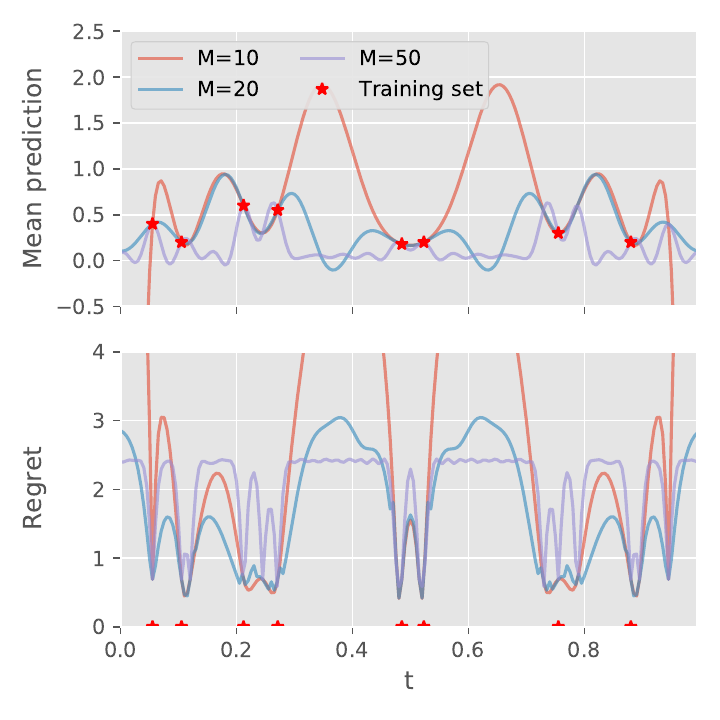}
    \vspace{-0.6cm}
    \caption{The pNML of different model degrees \label{fig:pnml_pred_and_regret}}
\end{subfigure}
\begin{subfigure}[t]{0.45\linewidth}
    \includegraphics[width=\columnwidth]{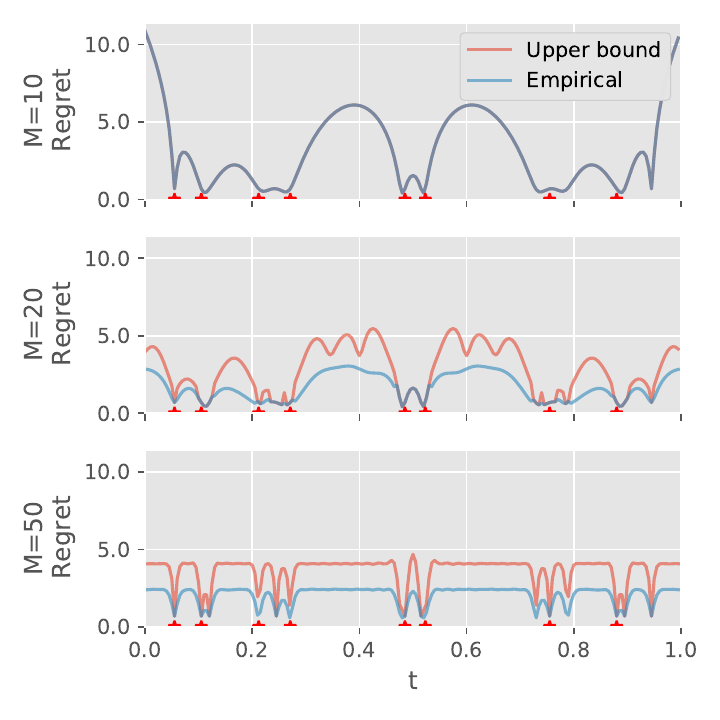}
    \vspace{-0.6cm}
    \caption{The pNML min-max regret \label{fig:pnml_syntetic_analytical_vs_empirical}}
\end{subfigure}
\caption{(Top left) The pNML prediction. (Bottom left) The pNML min-max regret. The training points are marked in red on the horizontal axis. (Right) The empirically calculated pNML regret and its analytical upper bound for models with a different number of learnable parameters.
The regret is low for test samples in the training data surroundings. More information in~\secref{sec:synthetic_data}.
}
\end{figure}

\subsection{Synthetic data} 
\label{sec:synthetic_data}

We use a training set that consists of 8 points $\{t_n,y_n\}_{n=0}^7$ in the interval $[0,1]$. These points are shown in \figref{fig:pnml_pred_and_regret} (top) as red dots. 
The data matrix was created with the following conversion:
\begin{equation}
X_{N}[n,m] = \cos \left(\pi m t_n +  \frac{\pi}{2} m\right) , \qquad
0 \leq n < N, \quad
0 \leq m <  M, 
\end{equation}
where $M$ and $N$ are the number of learnable parameters and the training set size respectively.
We predict using the pNML learner the labels of all $t$ values in the interval $[0,1]$.
\Figref{fig:pnml_pred_and_regret} (top) shows the mean pNML prediction for $M$ values of 10, 20, and 50. 
Since the number of parameters is greater than the training set size, all curves fit perfectly to the training points.

We treat each point in the interval $[0,1]$ as a test point and calculate its pNML min-max regret as shown in \figref{fig:pnml_pred_and_regret} (bottom).
The training $t_n$'s are marked in red on the horizontal axis. 
For every $M$, in the training data surroundings the regret is low comparing to areas where there are no training data. 

Surprisingly, the model with $M=10$ has a larger regret than models with a greater number of parameters. 
It may relate to the constraint: In this model, the norm value of the MN solution is 295,552  while for the models with $M=20$ and $M=50$ the MN solution norms are 0.15 and 0.04 respectively.
Having a lower norm constraint means a simpler model and better generalization.
This behaviour is also presented in the regret upper bound \eqref{eq:regret_upper_bound} with the term $\theta_N^{*\top} X_N^+ X_N^{+ \top} \theta_N^*$ that is proportional to the norm constraint value.
Furthermore, looking at $t=0.35$ for instance, the model with $M=10$ predicts a label that deviates from 0 much more than the model with $M=50$.

To show that the derived upper bound is informative we plot it along with the empirically calculated pNML min-max regret in~\figref{fig:pnml_syntetic_analytical_vs_empirical}.
For $M=10$ the analytical expression and the empirically calculated regret give the same results.
For the other model degrees, the upper bound and the empirical regret have a similar characteristic: In areas where the training data exists, the regret decreases, and as moving away to areas without training points, the regret increases.

\begin{figure}[bth]
\centering
\begin{subfigure}[t]{0.32\linewidth}
    \includegraphics[width=\textwidth]{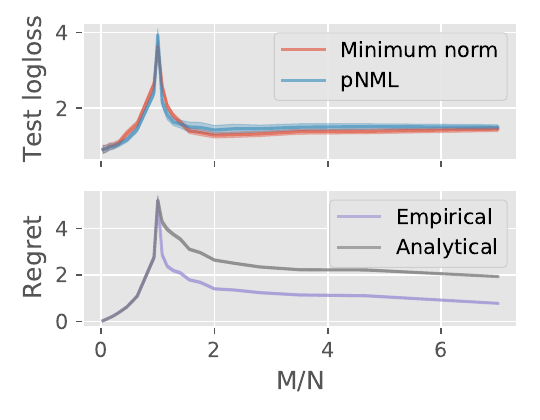}    
    \vspace{-0.6cm}
    \caption{Boston Housing \label{fig:bostonHousing}}
\end{subfigure}
\begin{subfigure}[t]{0.32\linewidth}
    \includegraphics[width=\textwidth]{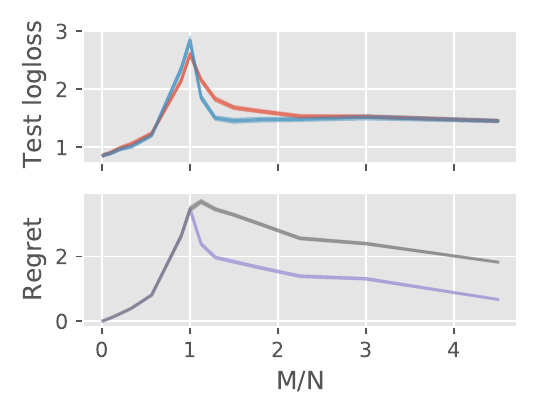}
    \vspace{-0.6cm}
    \caption{Concrete Compression Strength \label{fig:concrete}}
\end{subfigure}
\begin{subfigure}[t]{0.32\linewidth}
    \includegraphics[width=\textwidth]{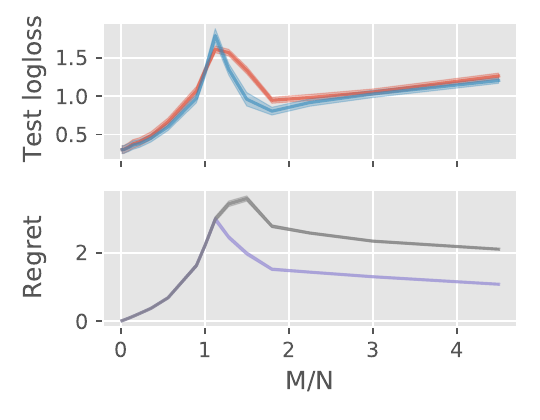}
    \vspace{-0.6cm}
    \caption{Energy Efficiency \label{fig:energy}}
\end{subfigure}
\begin{subfigure}[t]{0.32\linewidth}
    \includegraphics[width=\textwidth]{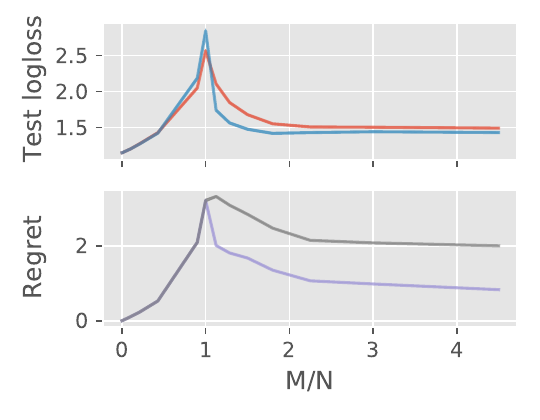}
    \vspace{-0.6cm}
    \caption{Kin8nm \label{fig:kin8nm}}
\end{subfigure}
\begin{subfigure}[t]{0.32\linewidth}
    \includegraphics[width=\textwidth]{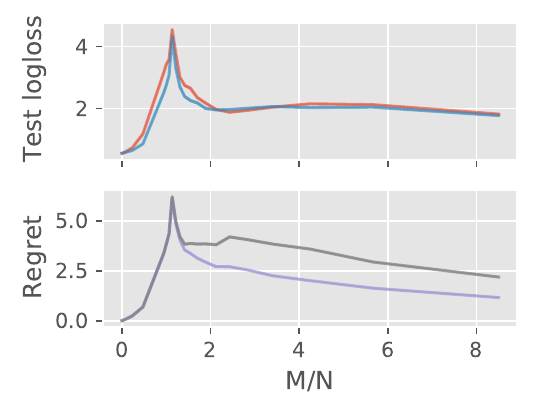}
    \vspace{-0.6cm}
    \caption{Naval Propulsion \label{fig:naval-propulsion-plant}}
\end{subfigure}
\begin{subfigure}[t]{0.32\linewidth}
    \includegraphics[width=\textwidth]{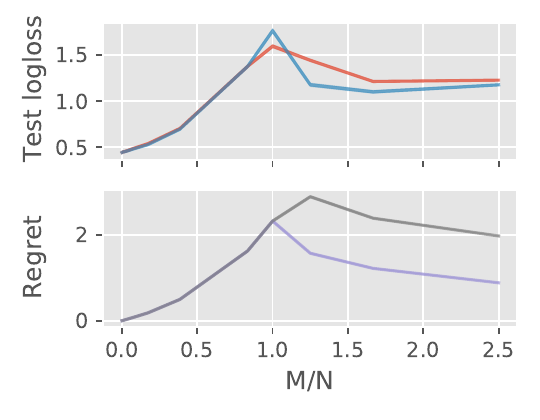}
    \vspace{-0.6cm}
    \caption{Combined Cycle Power Plant \label{fig:power-plant}}
\end{subfigure}
\begin{subfigure}[t]{0.32\linewidth}
    \includegraphics[width=\textwidth]{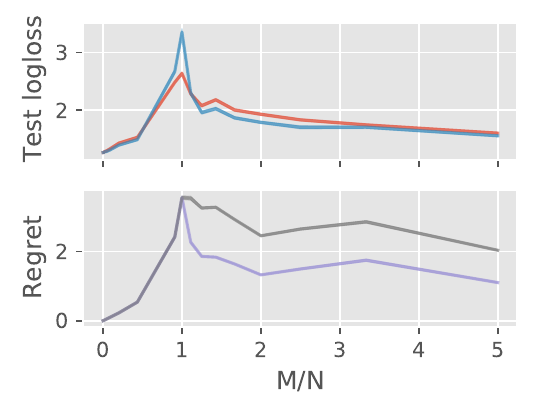}
    \vspace{-0.6cm}
    \caption{Protein Structure \label{fig:protein-structure}}
\end{subfigure}
\begin{subfigure}[t]{0.32\linewidth}
    \includegraphics[width=\textwidth]{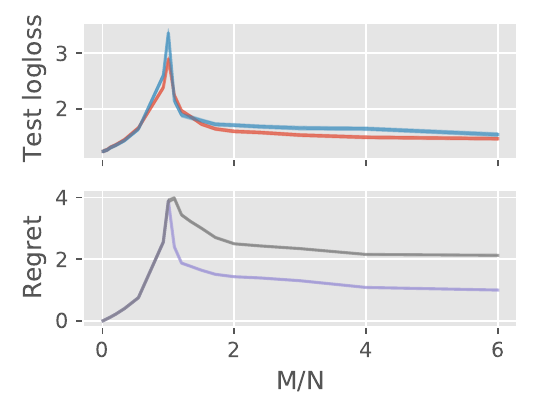}
    \vspace{-0.6cm}
    \caption{Wine Quality Red \label{fig:wine}}
\end{subfigure}
\begin{subfigure}[t]{0.32\linewidth}
    \includegraphics[width=\textwidth]{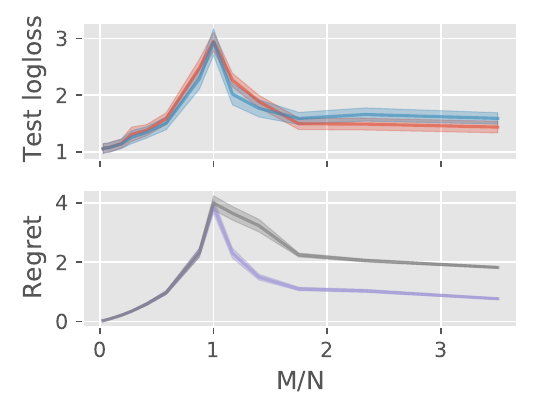}
    \vspace{-0.6cm}
    \caption{Yacht Hydrodynamics \label{fig:yacht}}
\end{subfigure}
\caption{The double-decent of the test logloss and pNML regret for UCI repository datasets with 95\% confidence interval. (Top) The test logloss for a varied number of  features ($M$) to trainset size ($N$) ratio. (Bottom) The empirically calculated regret and its upper bound. The regret presents the same double-descant behaviour as the test logloss. For more information see~\secref{sec:double_decent}.}
\label{fig:real_data_double_decent}
\end{figure}

\subsection{Double-decent with UCI dataset}
\label{sec:double_decent}

\begin{wraptable}{r}{0.475\linewidth}
\vspace{-0.4cm}
\small
\centering
\caption{UCI set characteristics}
\begin{tabular}{l c c c}
\toprule
Dataset name & $N$ & $M$ & \#Splits\\
\midrule
Boston Housing & 506 & 13 & 20 \\
Concrete Strength & 1,030 & 8 & 20 \\ 
Energy Efficiency & 768 & 8 & 20 \\
Kin8nm & 8,192 & 8 & 20 \\
Naval Propulsion & 11,934 & 16 & 20\\
Cycle Power Plant & 9,568 & 4 & 20\\ 
Protein Structure & 45,730 & 9  & 5 \\
Wine Quality Red & 1599 & 11 & 20\\
Yacht Hydrodynamics & 308 & 6 & 20\\
\bottomrule
\end{tabular}
\label{tab:uci_meta_data}
\vspace{-0.4cm}
\end{wraptable}

Double descent is referred to as the phenomenon when beyond the interpolation limit,  the test error declines as model complexity increases~\citep{hastie2019surprises}.
We investigate the effect of varying the ratio between the number of parameters to training set size. 
We demonstrate that the pNML regret and its upper bound are correlated with the double-decent behavior of the test logloss.

We utilize datasets from the UCI repository~\citep{Dua:2019}.
We use the sets proposed by~\citet{hernandez2015probabilistic} with the same test and train splits.
The training set size, number of features, and the number of train-test splits are shown in~\tableref{tab:uci_meta_data}.
For each dataset, we varied the training set size and fit the pNML and MN learners. We optimize $\sigma^2$ on a validation set for both learners.

The test set logloss as a function of the ratio between the number of parameters to training set size is presented in \figref{fig:real_data_double_decent} (top) with the 95\% confidence interval that was calculated on different train-test splits.
Both the pNML and the MN learners behave the same: For a large training set size ($\frac{M}{N}<1$) the test set logloss increases when removing training samples up to $M=N$. Then the logloss declines although the training set size decreases.

The empirically calculated pNML regret and its analytical upper bound are shown in \figref{fig:real_data_double_decent} (bottom).
Both empirically calculated regret and the derived upper bound present a similar double-decent behavior to the test logloss. Their peak is for the number of features that equals the training set size and as $\frac{M}{N}$ increases their value decrease.


\section{Conclusion} 
\label{sec:conclusion}

We derived an analytical upper bound of the pNML regret which is associated with the prediction uncertainty for over-parameterized linear regression.
The pNML prediction equals the MN solution thus the derived regret can be used to quantify the prediction uncertainty of the MN solution.
The derived result holds for a wide range of scenarios as we considered the individual setting where there is no assumption of a probabilistic relationship between the training and test.

Analyzing the pNML regret we can observe that if a test sample lies in the subspace spanned by the eigenvectors associated with large eigenvalues of the training data correlation matrix then over-parameterized linear regression generalizes well.
Finally, we provided simulations of the pNML for real trigonometric polynomial interpolation. We showed that the pNML regret can be used as a confidence measure and can is correlated with the test error double-decent phenomenon for 9 sets from the UCI repository.

\label{sec:limitations}
For future work, we would like to derive the explicit expression of the pNML regret rather than an upper bound. In addition, the pNML regret can be used for additional tasks such as active learning, probability calibration, and adversarial attack detection.

\nocite{bibas2021learning}
\FloatBarrier
\bibliographystyle{apalike}
\bibliography{main}

\appendix
\onecolumn
\title{On The Generalization of the Minimum Norm Solution for Over-parameterized Linear Regression
--Supplementary material--}

\section{The pNML for under-parameterized linear regression (Theorem 2)} \label{appendix:pnml_linear_regression}
Let $X_N \in R^{N \times M}$ as the matrix which contains all the training data and $Y_N \in R^{N}$ be label vector
\begin{equation}
X_{N} = \begin{bmatrix} x_1 & \dots & x_N \end{bmatrix}^\top, \quad
Y_{N} = \begin{bmatrix} y_1 & \dots & y_N \end{bmatrix}^\top.
\end{equation}
Given a test label with data $x$ and label $y$, the solution using the recursive least square formulation is
\begin{equation}
\thetagenie = \theta_{N} + \frac{\left(X_N^\top X_N \right)^{-1}}{1 + x^\top \left(X_N^\top X_N \right)^{-1} x}  x (y - x^\top  \theta_{N}),
\end{equation}
where $\theta_N$ is the least squares solution based on the $N$ training samples. The genie probability assignment is
\begin{equation}
\begin{split}
& p_{\thetagenie}(y|x) 
=
\frac{1}{\sqrt[]{2\pi\sigma^2}}\exp\left\{-\frac{1}{2\sigma^2}\left(y- x^\top \thetagenie \right)^2\right\} 
\\ & \qquad =
\frac{1}{\sqrt{2\pi\sigma^2}}\exp\bigg\{-\frac{1}{2\sigma^2}\left[y - x^\top  \left(\theta_{N} + \frac{\left(X_N^\top X_N \right)^{-1}}{1 + x^\top \left(X_N^\top X_N \right)^{-1} x} x (y - x^\top \theta_N) \right) \right]^2\bigg\} 
\\ & \qquad = 
\frac{1}{\sqrt[]{2\pi\sigma^2}}
\exp\left\{
-\frac{\left(y-x^\top \theta_N \right)^2}{2\sigma^2 \left[1 + x^\top \left(X_N^\top X_N\right)^{-1} x\right]^2} 
\right\}.  
\\
\end{split}
\end{equation}
To get the pNML normalization factor, we integrate over all possible labels
\begin{equation} \label{eq:underparam_norm_facotr}
K_0 = \int_{-\infty}^{\infty} \probthetagenietag dy'
= 1 + x^\top \left(X_N^\top X_N\right)^{-1} x.
\end{equation}
The pNML distribution of $y$ given the test data $x$ is
\begin{equation}
q_{\mbox{\tiny{pNML}}}(y |x) = \frac{1}{K}\probthetagenie 
= 
\frac{1}{\sqrt{2\pi\sigma^2} K_0}
\exp\left\{-\frac{\left(y - x^\top \theta_N \right)^2}{2\sigma^2 K_0^2}\right\} .
\end{equation}
The pNML regret, which is associate with the generalization error, is
\begin{equation} \label{eq:linear_regret}
\Gamma_0 = \log K_0 = \log\left[1 + x^\top \left(X_N^\top X_N\right)^{-1} x \right].
\end{equation}

\section{The norm of the minimum norm solution (Theorem 3)} \label{appendix:mn_norm_behaviour}
Let $c=x^\top \left(I - X_N^+ X_N \right)$.
For $c \neq 0$ the recursive formula to compute the pseudo-inverse of the data matrix is
\begin{equation}
X_{N+1}^+ = 
\begin{bmatrix} 
X_N^+ - c^+ x^\top X_N^+ & c^+ 
\end{bmatrix}.
\end{equation}
Denote the MN solution based on the $N$ training samples by $\theta_N^*$,
given a new sample $(x,y)$ the recursive formulation of the MN solution based on the training set and this new sample is
\begin{equation}
\theta_{N+1}^* 
= X_{N+1}^+ Y_{N+1} = 
\begin{bmatrix} 
X_N^+ - c^+ x^\top X_N^+ & c^+ 
\end{bmatrix}
\begin{bmatrix}
Y_N \\ y \\
\end{bmatrix} 
= 
\theta_N^* +  c^+ (y - x^\top \theta_N^*). 
\end{equation}
The norm of the MN solution based on these $N+1$ samples is
\begin{equation} \label{eq:mn_norm_recurisve}
||\theta_{N+1}^*||^2  
= 
||\theta_N^*||^2 + 2 \theta_N^{* \top} c^+ (y - x^\top \theta_N^*) + c^{+ ^\top} c^+ (y - x^\top \theta_N^*)^2.
\end{equation}

Denote $x_\bot = \left(I - X_N^+ X_N \right) x$, the pseudo-inverse of $c$ is
\begin{equation} \label{eq:c^+}
c^+ = c^\top (c c^\top)^{-1} 
= 
\left[x^\top \left(I - X_N^+ X_N \right)\right]^\top \frac{1}{x^\top \left[I -X_N^\top (X_N X_N^\top )^{-1} X_N \right] x} 
=
\frac{x_\bot}{||x_{\bot}||^2}.
\end{equation}
The inner product of the MN solution and $c^+$ can be written as
\begin{equation} \label{eq:theta_N_dot_c^+}
\theta_N^{* \top} c^+ 
=  
Y_N^\top X_N^{+ \top}
\frac{\left(I - X_N^+ X_N \right)^\top x}{x^\top \left(I - X_N^+ X_N \right)^2 x} 
= 
\frac{ Y_N^\top \left(X_N^+ - X_N^+\right)^\top x}{x^\top \left(I - X_N^+ X_N \right)^2 x} 
= 0.
\end{equation}

Substitute \eqref{eq:c^+} and \eqref{eq:theta_N_dot_c^+} to \eqref{eq:mn_norm_recurisve} gives the final result
\begin{equation} \label{eq:mn_norm_behaviour}
\norm{\theta_{N+1}}^2 = \norm{\theta_N^*}^2 + \frac{1}{\norm{x_{\bot}}^2} \left(y - x^\top \theta_N^* \right)^2.
\end{equation}

\section{The regularization factor lower bound (Lemma 2)}
\label{appendix:lambda_lower_bound}
We add the test sample $(x,y)$ to the training set.
The corresponding genie weights are
\begin{equation}
\thetagenie
=
\left(X_{N+1}^\top X_{N+1} + \lambda I \right)^{-1} X_{N+1}^\top
\begin{bmatrix} y_1 & \hdots & y_N & y \end{bmatrix}^\top.
\end{equation}
The MN least squares solution of the $N+1$ samples is $
\theta_{N+1}^*$.
Using Taylor series expansion of with respect to $\lambda$ and the MN recursive formulation
\begin{equation}
\begin{split}
\norm{\thetagenie}^2 
&\geq 
\norm{\theta_N^{* 2}}- 2 \theta_N^{*\top} X_{N+1}^+ X_{N+1}^{+ \top} \theta_N^{*} \lambda
\\ &=
\norm{\theta_N^{* 2}}- 2 \left( \theta_N^{* \top} X_N^+ X_N^{+ \top} \theta_N^*
+
\frac{\left(y - x^\top \theta_N^*\right)^2}{||x_{\bot}||^2} 
x^\top X_N^+ X_N^{+ \top} x  \right).
\end{split}
\end{equation}
The second equality is derived in \appref{appendix:taylor_series_second_term}. 
Utilizing \Theoref{theorem:mn_norm}
the following inequality is obtained
\begin{equation}
\begin{split}
\norm{\thetagenie}^2 
&\geq
\norm{\theta_N^*}^2 + \frac{1}{\norm{x_\bot}^2}\left(y - x^\top \theta_N^*\right)^2 
\\ & \qquad \qquad
- 2 \left(\theta_N^{* \top} X_N^+ X_N^{+ \top} \theta_N^*
+
\frac{\left(y - x^\top \theta_N^*\right)^2}{||x_{\bot}||^2} 
x^\top X_N^+ X_N^{+ \top} x \right)
\lambda.
\end{split}
\end{equation}
Plug it the norm constrain $\norm{\thetagenie}^2 = \norm{\theta_N^*}^2$:
\begin{equation} \label{eq:lambda_lower_bound}
\lambda \geq
\frac{1}{2}
\frac{\frac{1}{\norm{x_\bot}^2}\left(y - x^\top \theta_N^*\right)^2 }{\theta_N^{* \top} X_N^+ X_N^{+ \top} \theta_N^*
+
\frac{1}{||x_{\bot}||^2} 
x^\top X_N^+ X_N^{+ \top} x \left(y - x^\top \theta_N^*\right)^2}.
\end{equation}

\section{The pNML regret upper bound for over-parameterized linear regression (Theorem 4)}
\label{appendix:pnml_regret_upper_bound}
Denote $\delta \geq 0$ we relax the constraint
\begin{equation}
\norm{\theta_N^*}^2 = \norm{\theta_{N+1}}^2 \leq (1+\delta) \norm{\theta_N^*}^2.
\end{equation}
We get a perfect fit when to the following constraint is satisfied.
\begin{equation}
(1+\delta) \norm{\theta_N^*}^2
\geq 
\norm{\theta_N^*}^2 + \frac{1}{\norm{x_\bot}^2}\left(y' - x^\top \theta_N^* \right)^2
\end{equation}
\begin{equation}
x^\top \theta_N^* - \sqrt{\delta \norm{x_\bot}^2 \norm{\theta_N^*}^2} \leq y' \leq  x^\top \theta_N^* + \sqrt{\delta \norm{x_\bot}^2 \norm{\theta_N^*}^2}
\end{equation}

We split the integral of the normalization factor into two parts: one with a perfect fit and the other we upper bound with the genie upper bound (\Lemmaref{lemma:genie_upper_bound})
\begin{equation}
\begin{split}
K &\leq  
2 \int_{x^\top \theta_N^*}^{y^*} \frac{1}{\sqrt{2 \pi \sigma^2}} dy' +
2 \int_{y^*}^{\infty} 
\frac{1}{\sqrt{2\pi\sigma^2}}
\exp\left\{
- \frac{\left(y'-x^\top \theta_N \right)^2}{2\sigma^2 K_0^2 \left(1 + \frac{\norm{x_\bot}^2}{K_0 \lambda} \right)^2}
\right\}
dy'.
\\
&\leq  
2 \int_{x^\top \theta_N^*}^{y^*} \frac{1}{\sqrt{2 \pi \sigma^2}} dy' +
\int_{-\infty}^{\infty} 
\frac{1}{\sqrt{2\pi\sigma^2}}
\exp\left\{
- \frac{\left(y'-x^\top \theta_N \right)^2}{2\sigma^2 K_0^2 \left(1 + \frac{\norm{x_\bot}^2}{K_0 \lambda} \right)^2}
\right\}
dy'.
\\
&=
2 \sqrt{\frac{2 \delta}{\pi \sigma^2}  \norm{x_\bot}^2 \norm{\theta_N^*}^2}
+
K_0 \left(1 + \frac{\norm{x_\bot}^2}{K_0 \lambda}\right),
\end{split}
\end{equation}
where we fixed $\lambda$ by its lower bound \eqref{eq:lambda_lower_bound} at the point 
$y^*=x^\top \theta_N^* + \sqrt{\delta \norm{x_\bot}^2 \norm{\theta_N^*}^2}$. 
\begin{equation}
\begin{split}
K
&\leq  
2 \sqrt{\frac{2 \delta}{\pi \sigma^2}  \norm{x_\bot}^2 \norm{\theta_N^*}^2}
+
K_0 + 2 \norm{x_\bot}^2 \frac{\theta_N^{* \top} X_N^+ X_N^{+ \top} \theta_N^*
+
\delta \norm{\theta_N^*}^2
x^\top X_N^+ X_N^{+ \top} x }
{\delta \norm{\theta_N^*}^2}
\\ &=
\sqrt{\frac{2 \delta}{\pi \sigma^2}  \norm{x_\bot}^2 \norm{\theta_N^*}^2}
+
K_0 
+ 
2\norm{x_\bot}^2 x^\top X_N^+ X_N^{+ \top} x +
\frac{2 \norm{x_\bot}^2}{\delta \norm{\theta_N^*}^2}
\theta_N^{* \top} X_N^+ X_N^{+ \top} \theta_N^*
\\ &=
K_0 + 2\norm{x_\bot}^2 x^\top X_N^+ X_N^{+ \top} x
\\ & \qquad \qquad
+
\frac{2 \norm{x_\bot}^2}{\norm{\theta_N^*}^2}
\theta_N^{* \top} X_N^+ X_N^{+ \top} \theta_N^*
\left[
\frac{1}{\theta_N^{* \top} X_N^+ X_N^{+ \top} \theta_N^*}
\sqrt{\frac{\norm{\theta_N^*}^6}{2 \pi \sigma^2 \norm{x_\bot}^2}}
\sqrt{\delta}
+
\frac{1}{\delta}
\right].
\end{split}
\end{equation}
To find a tight upper bound, we choose $\delta$ that minimizes the right side
\begin{equation}
\begin{split}
K &\leq 
K_0 + 2\norm{x_\bot}^2 x^\top X_N^+ X_N^{+ \top} x
+
3\sqrt[3]{
\frac{1}{\pi \sigma^2}
\norm{x_\bot}^2
\theta_N^{* \top} X_N^+ X_N^{+ \top} \theta_N^*
}.
\end{split}
\end{equation}
Plugging in $K_0$, the theorem result is obtained.

\section{Taylor series second term} \label{appendix:taylor_series_second_term}
We prove that
\begin{equation}
\theta_N^{* \top} X_{N+1}^+ X_{N+1}^{+ \top} \theta_N^{*} =
\theta_N^{* \top} X_N^+ X_N^{+ \top} \theta_N^*
+
\frac{\left(y - x^\top \theta_N^*\right)^2}{||x_{\bot}||^2} 
x^\top X_N^+ X_N^{+ \top} x 
\end{equation}

We use the MN recursive formulation
\begin{equation} \label{eq:second_term_first_derv}
\begin{split}
& \theta_N^{* \top} X_{N+1}^+ X_{N+1}^{+ \top} \theta_N^{*}
=
\left(\theta_N^* + \frac{x_\bot}{||x_{\bot}||^2}  (y -x^\top \theta_N^* ) \right)^\top X_{N+1}^+ X_{N+1}^{+ \top} \left(\theta_N^* + \frac{x_\bot}{||x_{\bot}||^2}  (y -x^\top \theta_N^*) \right)
\\ & \quad = 
\theta_N^{* \top} X_{N+1}^+ X_{N+1}^{+ \top} \theta_N^* 
+
\frac{\left(y -x^\top \theta_N^*\right)^2}{||x_{\bot}||^2} 
x_\bot^\top X_{N+1}^+ X_{N+1}^{+ \top} x_\bot
+
\frac{2(y -x^\top \theta_N^*)}{||x_{\bot}||^2}  
x_\bot^\top X_{N+1}^+ X_{N+1}^{+ \top} \theta_N^*
\end{split}
\end{equation}
and
\begin{equation}
X_{N+1}^+ X_{N+1}^{+ \top}  
=
X_N^+ X_N^{+ \top} 
+
\frac{x^\top X_N^+
X_N^{+ \top}  x}{||x_{\bot}||^4} 
x_\bot x_\bot^\top
-
\frac{1}{||x_{\bot}||^2} 
X_N^+ X_N^{+ \top} x x_\bot^\top
-
\frac{1}{||x_{\bot}||^2} x_\bot x^\top X_N^+ X_N^{+ \top}.
\end{equation}

Substitute $X_{N+1}^+ X_{N+1}^{+ \top}$ to \eqref{eq:second_term_first_derv}:
\begin{equation}
\begin{split}
\theta_N^{* \top} X_{N+1}^+ X_{N+1}^{+ \top} \theta_N^{*}
&=
\theta_N^{* \top} X_N^+ X_N^{+ \top} \theta_N^* 
+
\frac{x^\top X_N^+ X_N^{+ \top}  x}{||x_{\bot}||^4}  \left(\theta_N^{* \top} x_\bot\right)^2
\\ &
-\frac{2}{||x_{\bot}||^2} x^\top X_N^+ X_N^{+ \top} \theta_N^* \left(\theta_N^{* \top} x_\bot \right)
\\ &
+
\frac{\left(y -x^\top \theta_N^*\right)^2}{||x_{\bot}||^2} 
\left[
x_\bot^\top
X_N^+ X_N^{+ \top} 
x_\bot
+
x^\top X_N^+ X_N^{+ \top}  x
-
2 x^\top X_N^+ X_N^{+ \top} x_\bot
\right]
\\ & 
+
\frac{2(y^* -x^\top \theta_N^*)}{||x_{\bot}||^2}
\bigg[
x_\bot^\top X_N^+ X_N^{+ \top} \theta_N^*
+
\frac{1}{||x_{\bot}||^2}
x^\top X_N^+ X_N^{+ \top}  x \left(\theta_N^{* \top} x_\bot \right)
\\ & \qquad \qquad \qquad 
-
\frac{1}{||x_{\bot}||^2} 
x_\bot^\top 
X_N^+ X_N^{+ \top} x \left(\theta_N^{* \top} x_\bot \right)
-
x_\bot^\top X_N^+ X_N^{+ \top} \theta_N^* 
\bigg].
\end{split}
\end{equation}
Most terms are zero since
\begin{equation}
x_\bot^\top \theta_N^* =  x^\top \left(I - X_N^\top X_N^{+ \top} \right) X_N^+ Y_N 
= Y_N^\top  \left(X_N^+ - X_N^+ \right)x = 0
\end{equation}
\begin{equation}
x_\bot^\top X_N^+ X_N^{+ \top} 
= x^\top \left(I - X_N^\top X_N^{+ \top} \right) X_N^+ X_N^{+ \top}
=  x^\top \left(X_N^+ X_N^{+ \top} - X_N^+ X_N^{+ \top} \right) 
= x^\top \mathbf{0}.
\end{equation}
And that proves the result.

\section{Real data: UCI dataset} \label{app:real_data_uci}
To evaluate the regret as a generalization measure we use datasets from the UCI repository as proposed by~\citet{hernandez2015probabilistic} with the same test and train splits.

For each dataset, we varied the training set size and fit the pNML and MN learners. We optimize $\sigma^2$ on a validation set for both learners.
We define a regret threshold and check the performance of the pNML taking into account only samples whose regret is lower than this threshold. We also evaluate the logloss of the MN learner of these samples. 

\begin{figure}[bt]
\centering
\begin{subfigure}[t]{0.315\linewidth}
    \includegraphics[width=1.0\textwidth]{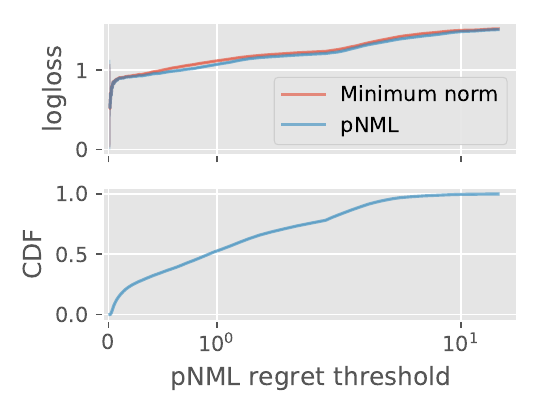}
    \vspace{-0.6cm}
    \caption{Boston Housing  \label{fig:bostonHousing_regret}}
\end{subfigure}
\begin{subfigure}[t]{0.315\linewidth}
    \includegraphics[width=1.0\textwidth]{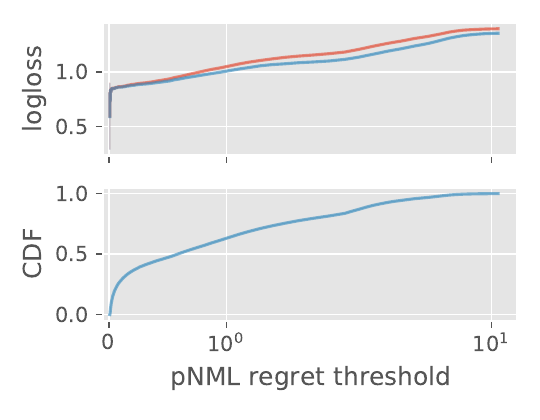}
    \vspace{-0.6cm}
    \caption{Concrete Compression Strength \label{fig:concrete_regret}}
\end{subfigure}
\begin{subfigure}[t]{0.315\linewidth}
    \includegraphics[width=1.0\textwidth]{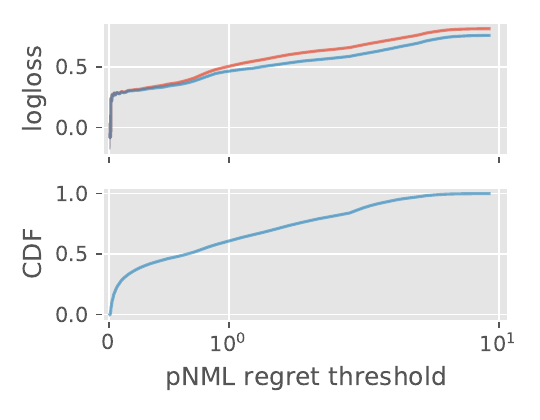}
    \vspace{-0.6cm}
    \caption{Energy Efficiency \label{fig:energy_regret}}
\end{subfigure}
\begin{subfigure}[t]{0.315\linewidth}
    \includegraphics[width=1.0\textwidth]{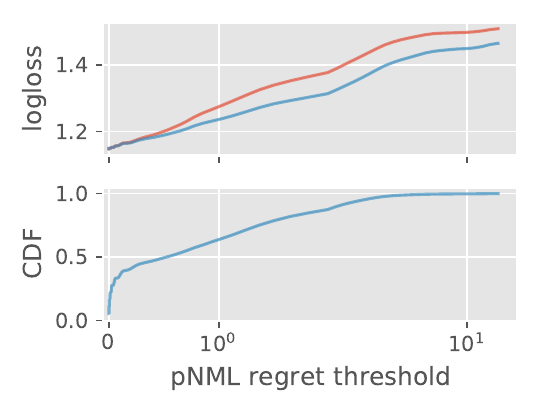}
    \vspace{-0.6cm}
    \caption{Kin8nm \label{fig:kin8nm_regret}}
\end{subfigure}
\begin{subfigure}[t]{0.315\linewidth}
    \includegraphics[width=1.0\textwidth]{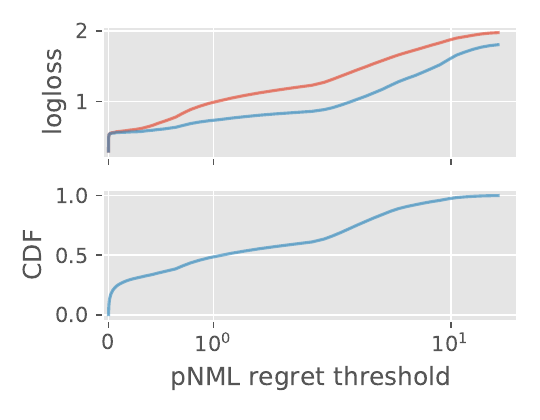}
    \vspace{-0.6cm}
    \caption{Naval Propulsion \label{fig:naval-propulsion-plant_regret}}
\end{subfigure}
\begin{subfigure}[t]{0.315\linewidth}
    \includegraphics[width=1.0\textwidth]{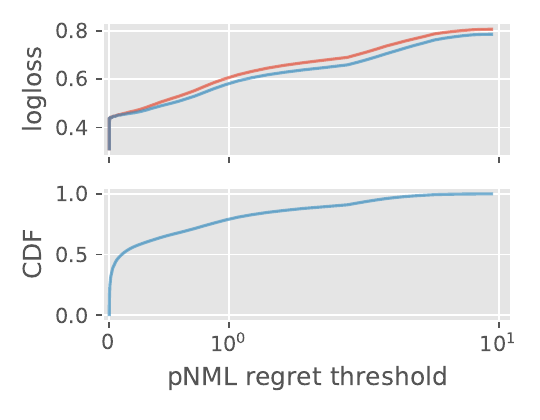}
    \vspace{-0.6cm}
    \caption{Combined Cycle Power Plant \label{fig:power-plant_regret}}
\end{subfigure}
\begin{subfigure}[t]{0.315\linewidth}
    \includegraphics[width=1.0\textwidth]{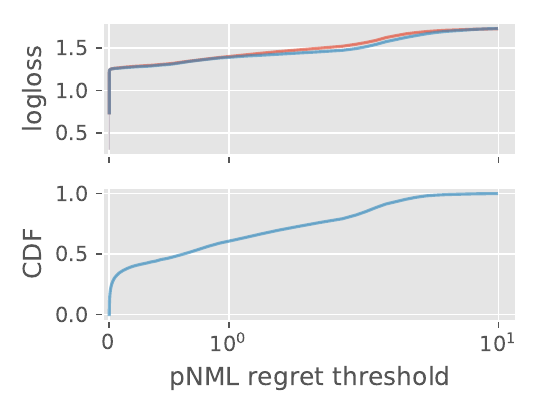}
    \vspace{-0.6cm}
    \caption{Protein Structure \label{fig:protein-structure_regret}}
\end{subfigure}
\begin{subfigure}[t]{0.315\linewidth}
    \includegraphics[width=1.0\textwidth]{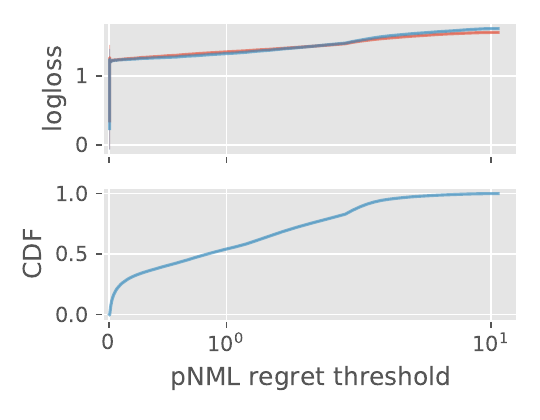}
    \vspace{-0.6cm}
    \caption{Wine Quality Red\label{fig:wine-quality-red_regret}}
\end{subfigure}
\begin{subfigure}[t]{0.315\linewidth}
    \includegraphics[width=1.0\textwidth]{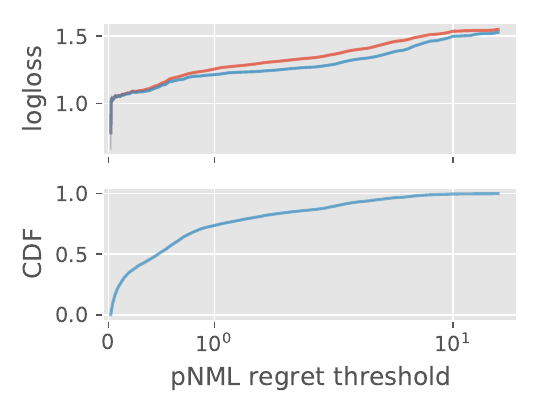}
    \vspace{-0.6cm}
    \caption{Yacht Hydrodynamics \label{fig:yacht_regret}}
\end{subfigure}
\caption{The performance of the MN solution and the pNML learner for test samples that have lower regret than the regret threshold. (Top) The test logloss of the learners as a function of the regret threshold. (Bottom) The CDF of the test samples.}
\label{fig:regret_based_learner}
\end{figure}

\Figref{fig:regret_based_learner} shows the logloss and the Cumulative Distribution Function (CDF) as function of the regret threshold with the 95\% confidence interval that was calculated on different train-test splits.
Both the test logloss of the pNML and MN learners are monotonically increasing functions of the regret threshold.
For 6 out of 10 datasets the pNML test logloss is lower than the MN while for the others the performance is equal.
Using the low regret as an indication for good generalization works the best in the Naval Propulsion dataset: the average test logloss of the 80\% of the samples with the lowest regret is 1.12, while the average logloss over all samples is 1.8.

\end{document}